\documentclass[11pt]{article}
\usepackage{acl}

\usepackage{times}
\usepackage{latexsym}
\usepackage[T1]{fontenc}
\usepackage[utf8]{inputenc}
\usepackage{microtype}
\microtypesetup{stretch=35,shrink=35,step=1}
\usepackage{inconsolata}
\usepackage{graphicx}
\usepackage{booktabs}
\usepackage{amsmath,amssymb}
\usepackage{multirow}
\usepackage{tabularx}
\usepackage{xcolor}
\usepackage{tikz}
\usetikzlibrary{arrows.meta, positioning, calc, fit, backgrounds}

\newcommand{\bu}{\mathbf{u}}
\newcommand{\bv}{\mathbf{v}}
\newcommand{\bphi}{\boldsymbol{\varphi}}
\newcommand{\cL}{\mathcal{L}}

\tikzset{
  blk/.style={
    draw=black!65, rounded corners=3pt,
    minimum height=0.62cm, text centered,
    font=\small, inner sep=5pt
  },
  wblk/.style={blk, fill=white},
  gblk/.style={blk, fill=gray!10},
  oblk/.style={blk, fill=orange!14},
  bblk/.style={blk, fill=blue!10},
  yblk/.style={blk, fill=yellow!14},
  nblk/.style={blk, fill=green!12},
  pblk/.style={blk, fill=purple!10},
  arr/.style={-{Stealth[length=5pt,width=4pt]}, thick, draw=black!65},
  darr/.style={arr, dashed, draw=gray!60},
}

\title{Building Interpretable Feature Representations for Resume--Vacancy\\
Matching by Distilling Production LLM Signals}

\author{Ilya Chekin\textsuperscript{1}, Vyacheslav Malyugin\textsuperscript{1},\\
Vladimir Chirkov\textsuperscript{1}, Mikhail Yurushkin\textsuperscript{2}\\
\textsuperscript{1}BroutonLab \quad \textsuperscript{2}Curately\\
\texttt{ilya.chekin@broutonlab.com, maliugin@broutonlab.com}\\
\texttt{vladimir.chirkov@broutonlab.com, mikhail.yurushkin@curatelyai.com}}

\hypersetup{
  pdftitle={Building Interpretable Feature Representations for Resume--Vacancy Matching by Distilling Production LLM Signals},
  pdfauthor={Ilya Chekin, Vyacheslav Malyugin, Vladimir Chirkov, Mikhail Yurushkin}
}

\begin{document}
\maketitle

\begin{abstract}
Matching candidates to vacancies is central to recruitment, and a
recruiter needs to see \emph{why} a candidate fits, not only a single
opaque relevance score. We provide this evidence as named,
interpretable matching dimensions recruiters can act on --- eight in
our current deployment.
We propose a two-part approach. The first is an LLM-based labeler whose
prompts and feature definitions were refined from recruiter feedback
while it served as an earlier production matching stage. In the current
architecture, it is used only for offline labeling and is not called on
online requests. The second is a feature
bi-encoder distilled from it: a LoRA-adapted embedding backbone with
compact per-dimension heads that runs on CPU and serves all online
requests. Both parts keep improving: prompts are revised as feedback
arrives, and the bi-encoder is retrained on the updated labels.
The model is trained on 168{,}772 labeled vacancy--resume pairs
(17{,}921 vacancies and 180{,}030 resumes).
Recruiters using the service can confirm or revise surfaced feature
predictions. On 927 recruiter-recorded values from this selected
production-feedback subset, the deployed student agrees with the
recorded decisions in 888 cases (\textbf{95.79\%}). This is operational,
non-blinded agreement rather than an independent human evaluation.
\end{abstract}

\section{Introduction}
\label{sec:intro}

Matching a candidate to a vacancy is not captured by a single number.
A candidate can fit an opening through job-title alignment, hard
skills, experience recency, seniority, certification requirements,
and other fine-grained dimensions.
A recruiter screening for an opening needs to know which of these
hold --- the named dimensions that explain why a candidate does or does
not fit --- and works with those values directly during screening.

A single relevance score collapses these dimensions into one value.
Such a score can order candidates, but it does not say why a candidate
fits: the recruiter cannot read off whether a low position reflects a
skill gap, a seniority mismatch, or a missing certification.
Person--job-fit systems that emit one fit score
\citep{Qin_2018,10.1145/3357384.3357949} are effective for ordering,
but on their own do not meet this requirement, since the final position
is not the signal a recruiter acts on.
We instead predict each matching dimension explicitly and surface it as
a named value.

To address this, we develop a two-part approach.
The first part uses an LLM that produces structured values for each
matching dimension of a candidate--vacancy comparison; its prompts and
feature definitions are refined from recruiter feedback
during production use, which lets us prototype complex matching logic
without collecting large manual label sets in advance.
This part enables fast adaptation of the logic but does not meet
production latency and serving-cost requirements
(\S\ref{sec:offline_engine}), so its role is to label data rather than to
be used in real business processes.

The second part is the distillation of the LLM labeler into a feature
bi-encoder.
Using the refined prompts, we annotate a large corpus of real
candidate--vacancy comparisons; on that corpus we train a cross-encoder
whose soft per-class distributions act as a teacher, and then a feature
bi-encoder whose representations expose the same interpretable
dimensions through compact heads (full pipeline in
Figure~\ref{fig:training}).
The distilled model serves on CPU and removes the per-resume
external API call while keeping the feature-level structure recruiters
rely on.
This is not a one-off migration: prompts continue to be revised as
feedback arrives, and the distilled model is retrained on the updated
labels, so the approach stays in an ongoing improvement loop. We
position our approach against the closest prior systems
in~\S\ref{sec:positioning}.

\section{Related Work}
\label{sec:related}

\paragraph{Recruitment and talent matching.}
Early person--job-fit systems learn a single match score over
structured vacancy and resume representations
\citep{Qin_2018,10.1145/3357384.3357949,bian2020learningmatchjobsresumes}, and
classical retrieval ranks candidates by embedding similarity
\citep{karpukhin2020densepassageretrievalopendomain,reimers2019sentencebertsentenceembeddingsusing}.
These systems return an order, not the reasons behind it, which is the
limitation our per-dimension output targets.
More recent work uses LLM-derived signals: \citet{li-etal-2025-enhancing-talent}
combine fine-grained LLM signals with a role-aware MoE ranker, and
several systems distill LLM matching signals into smaller scorers
\citep{jouanneau2026efficientlongcontextrankingarchitecture,fu2025lanternscalabledistillationlarge,yu2026confitv3improvingresumejob}.
These are the closest systems to ours; we compare against them in
detail in~\S\ref{sec:positioning} (Table~\ref{tab:related}).

\paragraph{Distillation and LLM annotation.}
Knowledge distillation \citep{hinton2015distillingknowledgeneuralnetwork,tang2019distillingtaskspecificknowledgebert}
and cross-encoder-to-bi-encoder distillation
\citep{hofstätter2021improvingefficientneuralranking,ren2023rocketqav2jointtrainingmethod} transfer a stronger
model's signal into an efficient one; cross-encoders
\citep{nogueira2020passagererankingbert} read both documents jointly, dense
bi-encoders \citep{karpukhin2020densepassageretrievalopendomain,reimers2019sentencebertsentenceembeddingsusing} encode them
independently. LLMs are increasingly used as annotation engines
\citep{Gilardi_2023}, and prior work compresses LLMs for
recommendation via distillation and pruning \citep{behdin-etal-2025-scaling};
cumulative-link ordinal models
\citep{mccullagh1980regression,cheng2007neuralnetworkapproachordinal} supply structural priors
for ordered labels. Augmented SBERT \citep{thakur2021augmentedsbert} also
transfers supervision from a cross-encoder to a bi-encoder, but uses the
cross-encoder to label additional sampled pairs and enlarge a smaller
training set. Here a production-refined LLM already supplies the hard
multidimensional labels, while the cross-encoder provides soft per-class
targets. We therefore treat cross-encoder-to-bi-encoder transfer as
established methodology; the system contribution is the feedback-refined
labeling policy, eight independently supervised dimensions,
stability-aware labeling, applicability gating, and the relabel/retrain
lifecycle. The detailed comparison with the closest systems appears
in~\S\ref{sec:positioning}.

\section{Industrial Setting and Task}
\label{sec:setting}

\subsection{System Evolution: Feedback-Driven Design}
\label{sec:feedback}

During production use, recruiters provided continuous
feedback on the surfaced feature values and failure cases.
This feedback was used to refine prompt instructions, clarify
feature definitions, and introduce new matching dimensions.
As a result, the final prompts are not ad-hoc LLM annotation
prompts but production-refined evaluation policies that reflect
operational recruiter requirements.

Comments frequently include per-feature corrections (e.g., ``Title
should be Match'') that directly informed prompt revisions; negative
ones indicate low-quality candidate sets.
A concrete example illustrates this evolution: observing that nominal
hard-skill overlap was insufficient when a skill had not been used
recently, we split the original ``skill match'' prompt into separate
\emph{skill context}, \emph{hard-skill experience}, \emph{skill
proficiency}, and \emph{experience recency} dimensions; a later cycle,
from regulated industries, added a \emph{certification} dimension with a
query-only applicability head (\S\ref{sec:applicability}) that activates
only when the vacancy requires a certification.

During the reported three-week pilot refinement period, the ML team
changed prompts manually; no second LLM or automatic prompt updater was
used. The process covered 247 human-reviewed feature values: 192 in a
fixed regression set and 55 newly observed error cases. Three prompt
updates were released, affecting Skill Context, Skill Proficiency, and
Certifications. A revision was accepted only if it corrected the target
failure pattern, generalized to related cases, produced an appropriate
explanation, and did not degrade the fixed regression set; the complete
protocol appears in Appendix~\ref{app:prompt_protocol}.

\subsection{Task and Production Constraints}
\label{sec:task}

Given a vacancy $j$ and a resume database $\mathcal{R} = \{r_1,
\ldots, r_n\}$, the approach matches each candidate against the
vacancy and outputs interpretable matching evidence.
We model fit as a tuple of $K{=}8$ matching dimensions
$\mathbf{f}(j,r) = (f_1(j,r), \ldots, f_K(j,r))$, with
$f_k \in \{0,1,2\}$ (ordinal) or $\{0,1\}$ (binary).
The eight dimensions --- job title match, industry experience,
certifications, hard-skill experience, skill proficiency, experience
recency, seniority alignment, and skill context --- are defined in full in
Appendix~\ref{app:features} (Table~\ref{tab:app_features}); seven are
ordinal (3 classes) and certification is binary.
Beyond multi-dimensional, asymmetric matching, the deployed system
must satisfy low latency, low cost, scalability to 150
candidates per vacancy, feature-level interpretability, and
robustness to heterogeneous document formats.

\subsection{Data}
\label{sec:data}

The approach is developed over a corpus drawn from a commercial
recruitment setting with recurring users across heterogeneous
domains and document formats.
The LLM-labeled training corpus contains 17{,}921 vacancies and
180{,}030 resumes, forming 168{,}772 labeled vacancy--resume pairs
across the eight matching dimensions; the split is by vacancy
(\S\ref{sec:offline_engine}).
Typically about 150 candidates fall to a vacancy, of which a recruiter
reviews a shortlist of about 50, and the system is in active
production~deployment.

\section{LLM-Distilled Feature Bi-Encoder}
\label{sec:model}

\subsection{Training Data Construction}
\label{sec:offline_engine}

\textbf{Cost and throughput.}
LLM labeling is unsuitable as a permanent online component: comparing
each candidate against the vacancy incurs a per-call monetary cost and
latency, and throughput is further capped by external API rate limits.
These costs scale linearly with traffic, so we distill the LLM into a
local model.

\textbf{The LLM stage as a labeling engine.}
After serving as the matching stage in an earlier production phase, the
LLM labeler now runs as an offline labeling engine rather than on online
requests: using the production-refined prompts, we annotate a large
corpus of comparisons with the same eight matching dimensions, and the
distilled bi-encoder serves those dimensions online. We do not treat the
labels as perfect ground truth but as production-refined supervision
signals. The prompt policy was refined through recruiter feedback and is
later checked against recruiter-recorded production decisions
(\S\ref{sec:human_eval}); that check is operational and non-blinded
because recruiters saw the deployed system prediction before confirming
or revising it. This process is not frozen: as new feedback arrives the
prompts are revised and the corpus is relabeled, which in turn feeds a
retraining of the distilled model.

\textbf{Run-to-run stability.}
Using the same OpenAI GPT-5.2 mini labeler, we generated each feature
five times on a stratified sample of 5{,}000 pairs. Across dimensions,
unanimous five-run agreement ranged from 86.7\% to 98.7\%, and the first
run matched the five-run majority in 94.8--99.7\% of cases. Six
dimensions achieved at least 98\% first-run/majority agreement and use
one run at scale. Industry Experience (95.9\%) and Seniority Alignment
(94.8\%) use adaptive majority voting: two runs initially, a third after
disagreement, and up to five for ambiguous cases. Only three examples
remained unstable: two corrupted or nearly empty resumes were excluded
and one ambiguous case went to manual review. This measures label
stability, not probabilistic calibration; further details appear in
Appendix~\ref{app:stability}.

\textbf{Data split and leakage sensitivity.}
The single train/validation split is vacancy-disjoint but was not
initially constrained to be candidate-disjoint. Of 180{,}030 unique
resumes, 148{,}555 occur in training and 31{,}539 in validation; only 64
candidates occur in both splits (0.04\% of training candidates and 0.20\%
of validation candidates). They always occur with different vacancies,
and no candidate--vacancy pair occurs in both splits. Removing these 64
candidates from validation changes accuracy/macro-F1 from 0.886/0.849 to
0.885/0.848, indicating no meaningful effect on the reported result.

\subsection{Encoder and Pair Representation}
\label{sec:encoder}

Unlike a standard retrieval bi-encoder, our model is trained so
that representations support extraction of interpretable matching
dimensions, not just a single similarity score.
We use a Qwen3-Embedding-0.6B backbone \citep{zhang2025qwen3embeddingadvancingtext} adapted
with LoRA \citep{hu2021loralowrankadaptationlarge} ($r{=}32$, $\alpha{=}64$).
Vacancy $j$ and resume $r$ are encoded independently into
1024-dim embeddings $E(\cdot)\in\mathbb{R}^{1024}$; resume embeddings
can be precomputed and cached.
Four components are concatenated into a pair representation:
\begin{equation}
\begin{split}
  \bphi(j,r) &= [E(j);\,E(r);\,|E(j){-}E(r)|; \\
             &\quad E(j){\odot}E(r)] \in \mathbb{R}^{4096}
\end{split}
  \label{eq:pair}
\end{equation}
The difference term captures dimensional misalignment and the
element-wise product captures agreement. This representation follows
established sentence-pair modeling practice
\citep{mou2016natural,conneau2017supervised,chen2017enhanced,reimers2019sentencebertsentenceembeddingsusing}
and is not claimed as novel.

\subsection{Task-Specific Heads}
\label{sec:heads}

\begin{figure*}[t]
\centering
\begin{tikzpicture}[
  node distance = 0.55cm and 0.8cm,
]
\node[gblk, minimum width=2.4cm] (q) at (-2.0, 3.2) {Vacancy $j$};
\node[oblk, minimum width=2.4cm] (eq) at (-2.0, 2.4) {$E(\cdot)$};
\node[gblk, minimum width=2.4cm] (u)  at (-2.0, 1.5) {$\bu\in\mathbb{R}^{1024}$};

\node[gblk, minimum width=2.4cm] (d)  at ( 2.0, 3.2) {Resume $r$};
\node[oblk, minimum width=2.4cm] (ed) at ( 2.0, 2.4) {$E(\cdot)$};
\node[gblk, minimum width=2.4cm] (v)  at ( 2.0, 1.5) {$\bv\in\mathbb{R}^{1024}$};

\draw[darr] (eq) -- node[above, font=\scriptsize, text=gray!75]{shared} (ed);

\node[yblk, minimum width=6.2cm, text width=6.0cm, align=center]
  (phi) at (0, 0.4) {$\bphi = [\bu;\bv;|\bu{-}\bv|;\bu{\odot}\bv] \;\in\mathbb{R}^{4096}$};

\node[gblk, minimum width=6.2cm]
  (heads) at (0, -0.5) {8 task-specific MLP heads $h_1,\ldots,h_{8}$};

\node[nblk, minimum width=6.2cm, text width=6.0cm, align=center]
  (out) at (0, -1.4) {Per-feature verdicts $\hat{y}_1,\ldots,\hat{y}_{8}$ {\small(7 ordinal, 1 binary)}};

\node[pblk, minimum width=2.9cm, text width=2.7cm, align=center]
  (gcert) at (-5.7, 0.4) {Applicability head (vacancy-only)\\ $g_{\mathrm{cert}}(E(j))$};
\node[pblk, minimum width=2.9cm, text width=2.7cm, align=center]
  (zcert) at (-5.7, -1.4) {$\hat{z}_{\mathrm{cert}}$: certification required?};

\draw[arr] (q)   -- (eq);
\draw[arr] (eq)  -- (u);
\draw[arr] (d)   -- (ed);
\draw[arr] (ed)  -- (v);
\draw[arr] (u)   -- (phi);
\draw[arr] (v)   -- (phi);
\draw[arr] (phi)   -- (heads);
\draw[arr] (heads) -- (out);

\draw[arr] (u.west) -| (gcert);
\draw[arr] (gcert) -- (zcert);
\draw[darr] (zcert.east) -- node[above, font=\scriptsize, text=gray!75]{masks} (out.west);
\end{tikzpicture}
\caption{Feature bi-encoder architecture. Vacancy and resume are
  encoded independently by a shared Qwen3-Embedding-0.6B + LoRA
  encoder. The four-component pair representation $\bphi$ feeds
  eight task-specific MLP heads that predict ordinal or binary
  matching features --- the interpretable per-feature verdicts shown
  to recruiters. A query-only certification applicability head
  (Eq.~\ref{eq:app_heads}, left) depends only on the vacancy embedding
  $E(j)$ and masks the certification dimension when it is not required
  by the vacancy.}
\label{fig:model}
\end{figure*}
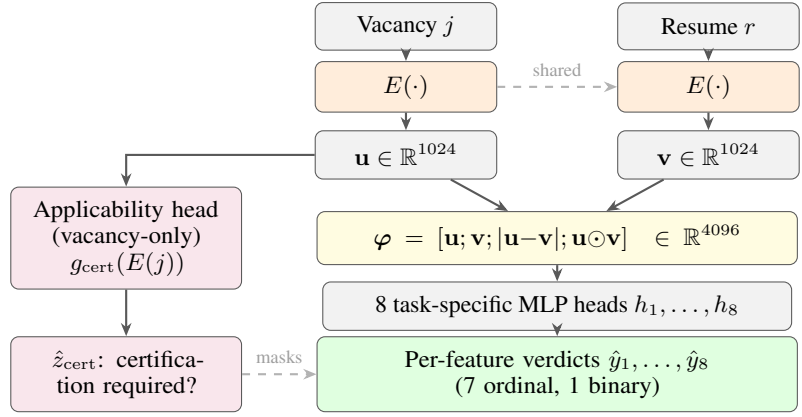

Each matching dimension is handled by an independent three-layer
MLP head $h_k: \mathbb{R}^{4096} \to \mathbb{R}^{o_k}$ with
LayerNorm, GELU, and dropout 0.1.
For the seven ordinal dimensions $o_k{=}2$ and for the one binary
dimension $o_k{=}1$ (Figure~\ref{fig:model}).
Separate heads preserve per-dimension interpretability and allow
independent evaluation, updating, or ablation of individual
features.

\paragraph{Ordinal modeling.}
For ordinal dimensions (those with ordered classes such as
old\,$<$\,recent\,$<$\,active), two output logits encode cumulative
probabilities via the proportional-odds model
\citep{mccullagh1980regression,cheng2007neuralnetworkapproachordinal} --- i.e., the model
predicts $P(\text{class} \ge t)$ for each threshold $t$, which
naturally preserves the label ordering.

\subsection{Query-Only Applicability Heads}
\label{sec:applicability}

Certification matching is meaningful only when the vacancy
explicitly requires a certification; applying it unconditionally
penalizes candidates for missing a requirement that was never
stated.
We train a lightweight binary applicability head on the vacancy
representation alone:
\begin{equation}
  \hat{z}_\text{cert} = g_\text{cert}(E(j))
  \label{eq:app_heads}
\end{equation}
Applicability is determined by the vacancy alone and is independent of
any particular resume, so we factor it into a separate head; adding a
``not applicable'' class directly to the pair MLP would conflate a
query-side signal with pairwise matching.
The head uses a two-layer trunk ($1024{\to}768{\to}384$, LN+ReLU)
followed by a linear classifier, trained jointly with the feature
heads via weighted BCE.
At inference, non-applicable certifications are excluded from the
surfaced output.

\section{Training Procedure}
\label{sec:training}

\subsection{Cross-Encoder Teacher}
\label{sec:ce}

Hard categorical labels discard relative mass among adjacent classes,
which limits the signal available for knowledge distillation.
We therefore train a Qwen3-4B reranker \citep{zhang2025qwen3embeddingadvancingtext} as an
intermediate cross-encoder on the LLM labels and store its
per-pair output logits $\mathbf{l}^{\mathrm{CE}}_k \in \mathbb{R}^3$
(ordinal) or $\mathbb{R}^1$ (binary) for use as a teacher during
bi-encoder training. Because the cross-encoder reads both documents
jointly, it acts as an intermediate soft-target teacher and produces a
richer training signal --- a per-class distribution rather than a single
chosen label.
On held-out pairs it reproduces the LLM labels at \textbf{0.934}
accuracy / \textbf{0.897} macro-F1 --- above the bi-encoder student
(0.886\,/\,0.849, Table~\ref{tab:feature_results}) --- so its soft
outputs carry information beyond the hard labels.

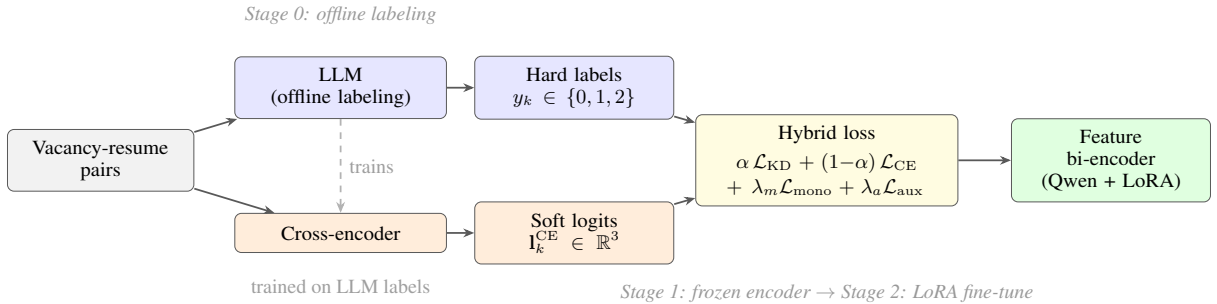
\begin{figure*}[t]
\centering
\resizebox{\textwidth}{!}{%
\begin{tikzpicture}[
  node distance = 0.6cm and 0.5cm,
]
\node[gblk, minimum width=2.8cm, text width=2.6cm, align=center]
  (pairs) at (0, 0) {Vacancy-resume\\pairs};
\node[bblk, minimum width=3.2cm, text width=3.0cm, align=center]
  (llm) at (3.8, 1.15) {LLM\\(offline labeling)};
\node[oblk, minimum width=3.2cm, text width=3.0cm, align=center]
  (ce)  at (3.8,-1.15) {Cross-encoder};
\node[font=\small, text=gray!75] at (3.8,-2.0) {trained on LLM labels};
\node[bblk, minimum width=3.0cm, text width=2.8cm, align=center]
  (hard) at (7.5, 1.15) {Hard labels\\$y_k\in\{0,1,2\}$};
\node[oblk, minimum width=3.0cm, text width=2.8cm, align=center]
  (soft) at (7.5,-1.15) {Soft logits\\$\mathbf{l}^{\mathrm{CE}}_k\in\mathbb{R}^3$};
\node[yblk, minimum width=4.0cm, text width=3.8cm, align=center]
  (loss) at (11.5, 0) {Hybrid loss\\[3pt]
    $\alpha\,\cL_{\mathrm{KD}} + (1{-}\alpha)\,\cL_{\mathrm{CE}}$\\[1pt]
    $+\;\lambda_m\cL_{\mathrm{mono}} + \lambda_a\cL_{\mathrm{aux}}$};
\node[nblk, minimum width=3.0cm, text width=2.8cm, align=center]
  (model) at (16.0, 0) {Feature\\bi-encoder\\(Qwen + LoRA)};

\draw[arr]  (pairs) -- (llm);
\draw[arr]  (pairs) -- (ce);
\draw[darr] (llm)   -- node[midway, right, font=\small, text=gray!75]{trains} (ce);
\draw[arr]  (llm)   -- (hard);
\draw[arr]  (ce)    -- (soft);
\draw[arr]  (hard)  -- (loss);
\draw[arr]  (soft)  -- (loss);
\draw[arr]  (loss)  -- (model);

\node[font=\small, text=gray!80, align=center] at (3.8, 2.3)
  {\textit{Stage 0: offline labeling}};
\node[font=\small, text=gray!80, align=center] at (11.5, -2.1)
  {\textit{Stage 1: frozen encoder $\to$ Stage 2: LoRA fine-tune}};
\end{tikzpicture}
}
\caption{Full training pipeline. Production-refined LLM prompts
  annotate vacancy-resume pairs offline with hard ordinal labels.
  A cross-encoder trained on these labels provides soft per-class
  targets for knowledge distillation. The feature
  bi-encoder is trained with a hybrid loss
  (Eq.~\ref{eq:loss}) combining KD, direct supervision,
  monotonicity, and margin regularization, in two stages
  (frozen encoder, then LoRA fine-tuning). The deployed student
  removes every online LLM call.}
\label{fig:training}
\end{figure*}

The cross-encoder is not used in production: it reads both documents
jointly and, at typical vacancy and resume lengths, requires a context
window of about 8{,}192 tokens with no cacheable representations, which
forces a GPU to stay online and does not meet our serving-cost targets.
The cross-encoder therefore serves only as an offline teacher, while
online requests are served by the distilled bi-encoder, which encodes
each document independently up to 4{,}096 tokens (the 4{,}096-token
limit covers ${>}99\%$ of vacancies and ${\approx}94\%$ of
resumes;~Table~\ref{tab:doc_lengths}).

\begin{table}[t]
  \centering
  \caption{Document length in tokens (training corpus).}
  \label{tab:doc_lengths}
  \small
  \resizebox{\columnwidth}{!}{%
  \begin{tabular}{lrrrrrr}
    \toprule
    \textbf{Document} & \textbf{n} & \textbf{p50} & \textbf{p75} & \textbf{p90} & \textbf{p95} & \textbf{p99} \\
    \midrule
    Vacancy & 17{,}921  & 503   & 725   & 997   & 1{,}236 & 2{,}018 \\
    Resume  & 180{,}030 & 1{,}321 & 2{,}038 & 3{,}235 & 4{,}205 & 6{,}244 \\
    \bottomrule
  \end{tabular}}
\end{table}

\subsection{Two-Stage Training}
\label{sec:stages}

Training proceeds in two stages: Stage~1 freezes the encoder and
trains only the MLP heads (stable initialization); Stage~2
unfreezes LoRA and jointly fine-tunes encoder and heads, adapting
the representation space to the matching-signal distribution.

\subsection{Hybrid Loss}
\label{sec:loss}

For each pair and task $k$, the total loss combines four terms:
\begin{equation}
  \cL = \alpha\,\cL_\text{KD}
      + (1{-}\alpha)\,\cL_\text{CE}
      + \lambda_m\,\cL_\text{mono}
      + \lambda_a\,\cL_\text{aux}
  \label{eq:loss}
\end{equation}
$\cL_\text{KD}$ matches student cumulative probabilities against
teacher cumulative probabilities (derived from the cross-encoder
logits) via BCE with temperature $T$.
$\cL_\text{CE}$ supervises against label-smoothed LLM hard labels.
$\cL_\text{mono}$ softly penalizes violations of the
non-decreasing constraint on cumulative probabilities (keeping the
ordinal predictions consistent).
$\cL_\text{aux}$ is a margin loss that pushes apart the
representations of pairs with different labels, with the margin
scaled by label distance.
Figure~\ref{fig:training} diagrams the full pipeline end to end.

\section{Feature Extraction and Interpretability}
\label{sec:pipeline}

A Qwen3-Embedding-0.6B retriever selects ${\sim}150$
candidates per vacancy via cosine similarity over cached embeddings
(geolocation and recency filters applied). The feature bi-encoder then
scores each candidate against the vacancy on all eight dimensions. Each
candidate is returned with a per-feature
class (e.g., title: match, skills: partial, certification: not
required), so the recruiter sees interpretable evidence for every
candidate rather than an opaque similarity number. The recruiter works
directly with named feature values --- the same dimensions they reason
about manually --- rather than only a fused similarity score. These are
directly supervised targets with fixed label spaces, not post-hoc
explanations of a global score: each dimension can be evaluated,
corrected, retrained, or masked independently. We focus evaluation on the ${\sim}50$
candidates a recruiter typically reviews per vacancy.

\section{Experiments}
\label{sec:experiments}

\subsection{Evaluation Setup}
\label{sec:eval_setup}

We report three distinct evaluations. \emph{Student-to-teacher fidelity}
measures how well distilled models reproduce LLM-derived labels on
held-out, vacancy-split corpus data (accuracy and macro-F1 over eight
dimensions; Appendix~\ref{app:per_feature}). \emph{Operational recruiter
agreement} evaluates the deployed student and, separately, the original
LLM labeler against recruiter-recorded production decisions on the
available feedback subset. Recruiters saw the deployed system prediction
before recording a value and could confirm or revise it; the original
labeler was compared separately with those same recorded values. Only
selected cases and features were recorded, so this is non-blinded
operational agreement, not independent human accuracy. \emph{Efficiency}
measures serving throughput and cost.

\subsection{Student-to-Teacher Fidelity and Ablations}
\label{sec:feature_results}

The cross-encoder provides the highest fidelity but requires joint
long-context GPU inference and cannot cache resume representations. The
bi-encoder trades some fidelity for cacheable embeddings and CPU serving.
Compared with the original frozen encoder, adapting the representation
and heads raises macro-F1 from 0.669 to 0.849; the controlled ablations
below isolate the pair representation and loss terms rather than
attributing that combined gain to one component. Seniority Alignment has
the lowest F1 (0.795), likely because seniority is recruiter-specific
(one recruiter's ``senior'' is another's ``mid-level''), making it a
candidate for future decomposition.

\begin{table}[t]
  \centering
  \caption{Fidelity to LLM labels on the single held-out,
  vacancy-disjoint corpus split described in \S\ref{sec:offline_engine},
  averaged over eight dimensions; these are not human ground-truth
  metrics. The cross-encoder is the offline teacher and the bi-encoder
  is the production student. Bold: best fidelity.}
  \label{tab:feature_results}
  \small
  \begin{tabular}{lcc}
    \toprule
    \textbf{Method} & \textbf{Acc.} & \textbf{Macro-F1} \\
    \midrule
    Majority baseline       & 0.606 & 0.268 \\
    Original encoder + MLP  & 0.760 & 0.669 \\
    Cross-encoder (offline) & \textbf{0.934} & \textbf{0.897} \\
    Bi-encoder (production) & 0.886 & 0.849 \\
    \bottomrule
  \end{tabular}
\end{table}

The absolute-difference term contributes more individually than the
element-wise product, while the product adds complementary signal.
Knowledge distillation provides the largest individual gain over
hard-label cross-entropy; ordinal consistency and margin regularization
provide smaller complementary improvements.

\begin{table*}[t]
\centering\small
\setlength{\tabcolsep}{6pt}
\begin{tabular}{@{}lc@{\qquad}lc@{}}
\toprule
\multicolumn{2}{c}{\textbf{(a) Pair representation}} &
\multicolumn{2}{c}{\textbf{(b) Training objective}} \\
\midrule
$[\bu;\bv]$ & 0.837 & Hard-label cross-entropy & 0.825 \\
$[\bu;\bv;\bu\odot\bv]$ & 0.842 & + cross-encoder soft targets (KD) & 0.840 \\
$[\bu;\bv;|\bu-\bv|]$ & 0.846 & + KD + ordinal consistency & 0.845 \\
$[\bu;\bv;|\bu-\bv|;\bu\odot\bv]$ & \textbf{0.849} &
+ KD + margin (without ordinal) & 0.844 \\
& & Full objective & \textbf{0.849} \\
\bottomrule
\end{tabular}
\caption{Controlled macro-F1 ablations. The pairwise terms follow
established sentence-pair practice; the full representation and full
objective perform best.}
\label{tab:ablations}
\end{table*}

\subsection{Operational Agreement with Recruiter-Recorded Decisions}
\label{sec:human_eval}

The feedback is a by-product of normal service use rather than a
separate labeling task. Recruiters saw the deployed system prediction
before recording a value and could confirm or revise it. Feedback was
provided only for selected candidates and features; we therefore report
exact agreement with recruiter-recorded production decisions, not
accuracy against an independent blinded reference. Of 873 reviewed
candidates, the correction form was opened for 144; 24 forms were empty
and 120 recorded at least one value, corresponding to 114 unique
candidates and 45 vacancies. Among them, 87 included the later-added
Experience Recency dimension, yielding 927 recorded feature values.

\textbf{LLM-to-recruiter operational agreement.}
The original OpenAI GPT-5.2 mini labeler, using the prompt version that
constructed the corpus, agrees with the 927 recorded decisions on
924/927 values (99.7\%; macro-F1 99.6\%). Seven dimensions have 100\%
observed agreement on this subset. All three disagreements are in
Seniority Alignment: recruiters selected aligned while the LLM selected
the more conservative partial; that dimension has 97.5\% agreement and
97.2\% macro-F1. This checks operational consistency on the available
selected feedback, not blinded human accuracy or calibrated uncertainty.

\textbf{Deployed-student-to-recruiter operational agreement.}
The deployed student exactly matches 888/927 recorded values (95.79\%;
Wilson 95\% CI 94.30--96.91\%). Candidate-cluster and vacancy-cluster
bootstraps (10{,}000 iterations) give 95\% CIs of 94.67--96.89\% and
94.11--97.37\%, respectively (Appendix~\ref{app:stability}). Seniority
Alignment is the weak dimension at 86/120 (71.7\%; Wilson CI
63.03--78.96\%); the other dimensions have much higher agreement. The 39
disagreements comprise 34 Seniority Alignment, three Certifications, and
two Job Title Match cases.

\begin{table*}[t]
  \centering\small
  \setlength{\tabcolsep}{5pt}
  \begin{tabular}{lrrrrr}
    \toprule
    \textbf{Dimension} & \textbf{n} & \textbf{Exact} & \textbf{Agree} &
      \textbf{Wilson 95\% CI} & \textbf{Maj.} \\
    \midrule
    Skill Context         & 120 & 120 & 100.0\% & 96.90--100.00\% & 72.5\% \\
    Skill Proficiency     & 120 & 120 & 100.0\% & 96.90--100.00\% & 61.7\% \\
    Hard Skill Experience & 120 & 120 & 100.0\% & 96.90--100.00\% & 38.3\% \\
    Industry Experience   & 120 & 120 & 100.0\% & 96.90--100.00\% & 47.5\% \\
    Experience Recency    &  87 &  87 & 100.0\% & 95.77--100.00\% & 87.4\% \\
    Job Title Match       & 120 & 118 &  98.3\% & 94.13--99.54\% & 35.8\% \\
    Certifications        & 120 & 117 &  97.5\% & 92.91--99.15\% & 83.3\% \\
    Seniority Alignment   & 120 &  86 &  71.7\% & 63.03--78.96\% & 68.3\% \\
    \midrule
    \textbf{Overall}     & \textbf{927} & \textbf{888} & \textbf{95.79\%} &
       \textbf{94.30--96.91\%} & \textbf{60.9\%} \\
    \bottomrule
  \end{tabular}
  \caption{Agreement of the deployed student with recruiter-recorded
  production decisions. Recruiters saw the deployed system prediction
  before confirming or revising it; the analysis is non-blinded and uses
  only explicitly recorded feedback. Wilson intervals quantify
  uncertainty in the observed proportions; 100\% is sample agreement,
  not perfect population accuracy.}
  \label{tab:feedback}
\end{table*}

The 0.849 macro-F1 and 95.79\% agreement use different datasets, targets,
and metrics. The former is student-to-LLM fidelity on the held-out corpus
and weights classes equally; the latter is exact agreement with
recruiter-recorded values on a small, selected feedback subset with a
different class distribution.

\subsection{Efficiency}
\label{sec:efficiency}

The deployed model has low operational cost. Resume embeddings are
precomputed offline (on a T4) and cached, so the only online cost is the
lightweight MLP heads: about 5{,}500 resumes/s on a single CPU with no
GPU or external API access, corresponding to roughly
\$$4.7\times10^{-5}$ per resume including the offline precompute.
Serving is fully async, and end-to-end latency is dominated by a fixed
network round-trip ($\le$500\,ms); heavy online LLM inference is removed
from the pipeline entirely.

\section{Positioning Relative to Closest Work}
\label{sec:positioning}

The closest line distills LLM matching signals into smaller
person--job-fit scorers: \citet{jouanneau2026efficientlongcontextrankingarchitecture} use a
generative LLM teacher with a calibrated distillation loss for
interpretable skill-fit scores; \citet{fu2025lanternscalabledistillationlarge} distill LLMs for
job--person fit with natural-language explanations; and
\citet{yu2026confitv3improvingresumejob} improve resume--job matching with LLM
re-ranking.
Our setting differs in three ways (summarized in
Table~\ref{tab:related}):
(i) supervision is a \emph{production-refined}
policy stabilized through recruiter feedback while the LLM served as the
matching stage in an earlier deployment --- the continued development of an
already-shipped approach, not a distillation built from scratch; (ii) the
student exposes eight independently supervised,
named ordinal dimensions, not a single skill-fit score
or free-text rationale; (iii) it runs on CPU and is checked against
available recruiter-recorded feedback on surfaced per-dimension outputs
(\S\ref{sec:human_eval}); the full comparison is given in
Appendix~\ref{app:related} (Table~\ref{tab:related}).
Table~\ref{tab:related} is qualitative rather than a direct quantitative
benchmark. The closest systems use different, mostly proprietary datasets
and predict a single relevance/skill-fit score or free-text explanation
rather than our eight categorical dimensions; their published F1/accuracy
values are therefore not directly comparable. We instead report
controlled internal baselines on our corpus (\S\ref{sec:feature_results}).

\section{Production Lessons}
\label{sec:lessons}

\textbf{(1)}~LLMs bootstrap matching logic without manual labels, and
recruiter feedback on the live system is used to refine the supervision
signal.
\textbf{(2)}~Per-dimension outputs localize errors (skills, seniority,
recency, certification) where a single fused score cannot.
\textbf{(3)}~Within one approach the LLM is best used as an offline
labeling engine that keeps adapting to feedback, while a distilled model
serves requests online.
\textbf{(4)}~Not every feature applies to every vacancy: a requirement
like certification is relevant only if the vacancy states it.
Query-activated applicability heads score such a feature only when the
vacancy presents it, and do not penalize a candidate for failing a
requirement the vacancy never~raised.

\section{Conclusion}
\label{sec:conclusion}

We presented an approach for automated candidate--vacancy matching over
named matching dimensions (eight in the current deployment). A
feedback-refined LLM labeler labels comparisons, and a distilled feature
bi-encoder serves those dimensions on CPU without per-comparison API
calls, so recruiters see named per-dimension values rather than an opaque
score. On the available non-blinded production-feedback subset, the
deployed student agrees with 888/927 recruiter-recorded values
(95.79\%; Wilson 95\% CI 94.30--96.91\%).

\section*{Limitations}

The LLM prompt policy was refined through recruiter feedback and
operationally checked against recruiter-recorded decisions, but this
feedback is sparse, selected, and non-blinded. It may be delayed or
mistaken, and each recorded value comes from one recruiter, so we cannot
report inter-annotator agreement; multiply judged values are future work.

The dataset we use contains proprietary vacancy and resume data and
cannot be publicly released, limiting external reproducibility.
Although the corpus covers multiple domains and document formats,
further evaluation is needed on entirely new labor-market segments.
We lack a directly comparable external quantitative benchmark because
the closest published systems use different, largely proprietary data
and different output targets. We also did not measure screening time,
shortlist quality, time-to-hire, or other downstream hiring outcomes, so
we do not claim causal business impact; controlled measurement of these
outcomes is future work.
Education level and language requirements were identified as
relevant matching dimensions through recruiter feedback but are
not included in the current system due to insufficient labeled
data for reliable training; incorporating them is left for future
work.
In addition, the system is trained and evaluated only on
English-language vacancies and resumes; transfer to other languages is
left for future work.
The approach is intended to support recruiters, not to make
autonomous hiring decisions.

\section*{Ethical Considerations}

Recruitment matching systems can affect candidate opportunities.
We frame the proposed approach as decision support
rather than an autonomous hiring mechanism.
The model should not use protected attributes as matching
features, and organizations deploying such systems should monitor
outputs for potential disparate impact.
The source-platform owner anonymizes the data before it reaches our
systems: names, contact details, identifiers, and explicitly stated
protected attributes are removed, and we cannot access the original
versions. This does not guarantee the absence of bias: indirect textual
proxies may remain, and bias in the foundation LLM used for labeling may
be transferred through distillation. We did not conduct a dedicated
empirical bias audit in this study. Counterfactual testing and
group-disparity analysis are necessary future work before stronger
fairness claims can be made.
Feature-level explanations should be interpreted as model predictions,
not objective assessments of candidate ability; the system is intended
for recruiter decision support rather than autonomous hiring, and human
oversight is required before consequential decisions.

\bibliography{references}

@inproceedings{li-etal-2025-enhancing-talent,
  title     = {Enhancing Talent Search Ranking with Role-Aware Expert Mixtures and {LLM}-based Fine-Grained Job Descriptions},
  author    = {Li, Jihang  and
               Xu, Bing  and
               Chen, Zulong  and
               Xu, Chuanfei  and
               Chen, Minping  and
               Liu, Suyu  and
               Zhou, Ying  and
               Wen, Zeyi},
  editor    = {Potdar, Saloni  and
               Rojas-Barahona, Lina  and
               Montella, Sebastien},
  booktitle = {Proceedings of the 2025 Conference on Empirical Methods in Natural Language Processing: Industry Track},
  month     = nov,
  year      = {2025},
  address   = {Suzhou (China)},
  publisher = {Association for Computational Linguistics},
  url       = {https://aclanthology.org/2025.emnlp-industry.13/},
  doi       = {10.18653/v1/2025.emnlp-industry.13},
  pages     = {185--194},
  isbn      = {979-8-89176-333-3}
}

@inproceedings{behdin-etal-2025-scaling,
  title     = {Scaling Down, Serving Fast: Compressing and Deploying Efficient {LLM}s for Recommendation Systems},
  author    = {Behdin, Kayhan  and
               Fatahibaarzi, Ata  and
               Song, Qingquan  and
               Dai, Yun  and
               Gupta, Aman  and
               Wang, Zhipeng  and
               Sang, Hejian  and
               Tang, Shao  and
               Dexter, Gregory  and
               Zhu, Sirou  and
               Zhu, Siyu  and
               Dharamsi, Tejas  and
               Kothapalli, Vignesh  and
               Fu, Zhoutong  and
               Cao, Yihan  and
               Hsu, Pin-Lun  and
               Borisyuk, Fedor  and
               Pillai, Natesh S.  and
               Simon, Luke  and
               Mazumder, Rahul},
  editor    = {Potdar, Saloni  and
               Rojas-Barahona, Lina  and
               Montella, Sebastien},
  booktitle = {Proceedings of the 2025 Conference on Empirical Methods in Natural Language Processing: Industry Track},
  month     = nov,
  year      = {2025},
  address   = {Suzhou (China)},
  publisher = {Association for Computational Linguistics},
  url       = {https://aclanthology.org/2025.emnlp-industry.119/},
  doi       = {10.18653/v1/2025.emnlp-industry.119},
  pages     = {1687--1702},
  isbn      = {979-8-89176-333-3}
}

@misc{zhang2025qwen3embeddingadvancingtext,
  title         = {Qwen3 Embedding: Advancing Text Embedding and Reranking Through Foundation Models},
  author        = {Yanzhao Zhang and Mingxin Li and Dingkun Long and Xin Zhang and Huan Lin and Baosong Yang and Pengjun Xie and An Yang and Dayiheng Liu and Junyang Lin and Fei Huang and Jingren Zhou},
  year          = {2025},
  eprint        = {2506.05176},
  archiveprefix = {arXiv},
  primaryclass  = {cs.CL},
  url           = {https://arxiv.org/abs/2506.05176}
}

@inproceedings{hu2021loralowrankadaptationlarge,
  title={{LoRA}: {Low-Rank} Adaptation of Large Language Models},
  author={Hu, Edward J. and Shen, Yelong and Wallis, Phillip and Allen-Zhu, Zeyuan and Li, Yuanzhi and Wang, Shean and Wang, Lu and Chen, Weizhu},
  booktitle={International Conference on Learning Representations},
  year={2022}
}

@misc{nogueira2020passagererankingbert,
  title         = {Passage Re-ranking with BERT},
  author        = {Rodrigo Nogueira and Kyunghyun Cho},
  year          = {2020},
  eprint        = {1901.04085},
  archiveprefix = {arXiv},
  primaryclass  = {cs.IR},
  url           = {https://arxiv.org/abs/1901.04085}
}

@misc{hofstätter2021improvingefficientneuralranking,
  title         = {Improving Efficient Neural Ranking Models with Cross-Architecture Knowledge Distillation},
  author        = {Sebastian Hofstätter and Sophia Althammer and Michael Schröder and Mete Sertkan and Allan Hanbury},
  year          = {2021},
  eprint        = {2010.02666},
  archiveprefix = {arXiv},
  primaryclass  = {cs.IR},
  url           = {https://arxiv.org/abs/2010.02666}
}

@inproceedings{ren2023rocketqav2jointtrainingmethod,
  title={{RocketQAv2}: A Joint Training Method for Dense Passage Retrieval and Passage Re-ranking},
  author={Ren, Ruiyang and Qu, Yingqi and Liu, Jing and Zhao, Wayne Xin and She, QiaoQiao and Wu, Hua and Wang, Haifeng and Wen, Ji-Rong},
  booktitle={Proceedings of the 2021 Conference on Empirical Methods in Natural Language Processing},
  pages={2825--2835},
  address={Online and Punta Cana, Dominican Republic},
  publisher={Association for Computational Linguistics},
  year={2021}
}

@article{mccullagh1980regression,
  author  = {McCullagh, Peter},
  title   = {Regression Models for Ordinal Data},
  journal = {Journal of the Royal Statistical Society: Series B (Methodological)},
  year    = {1980},
  volume  = {42},
  number  = {2},
  pages   = {109--127},
  doi     = {10.1111/j.2517-6161.1980.tb01109.x},
  url     = {https://doi.org/10.1111/j.2517-6161.1980.tb01109.x}
}

@inproceedings{cheng2007neuralnetworkapproachordinal,
  title={A Neural Network Approach to Ordinal Regression},
  author={Cheng, Jianlin and Wang, Zheng and Pollastri, Gianluca},
  booktitle={2008 IEEE International Joint Conference on Neural Networks (IEEE World Congress on Computational Intelligence)},
  pages={1279--1284},
  publisher={IEEE},
  year={2008}
}

@inproceedings{reimers2019sentencebertsentenceembeddingsusing,
  title={Sentence-{BERT}: Sentence Embeddings using {Siamese} {BERT}-Networks},
  author={Reimers, Nils and Gurevych, Iryna},
  booktitle={Proceedings of the 2019 Conference on Empirical Methods in Natural Language Processing and the 9th International Joint Conference on Natural Language Processing (EMNLP-IJCNLP)},
  pages={3982--3992},
  address={Hong Kong, China},
  publisher={Association for Computational Linguistics},
  year={2019}
}

@inproceedings{karpukhin2020densepassageretrievalopendomain,
  title={Dense Passage Retrieval for Open-Domain Question Answering},
  author={Karpukhin, Vladimir and O\u{g}uz, Barlas and Min, Sewon and Lewis, Patrick and Wu, Ledell and Edunov, Sergey and Chen, Danqi and Yih, Wen-tau},
  booktitle={Proceedings of the 2020 Conference on Empirical Methods in Natural Language Processing (EMNLP)},
  pages={6769--6781},
  address={Online},
  publisher={Association for Computational Linguistics},
  year={2020}
}

@misc{hinton2015distillingknowledgeneuralnetwork,
  title         = {Distilling the Knowledge in a Neural Network},
  author        = {Geoffrey Hinton and Oriol Vinyals and Jeff Dean},
  year          = {2015},
  eprint        = {1503.02531},
  archiveprefix = {arXiv},
  primaryclass  = {stat.ML},
  url           = {https://arxiv.org/abs/1503.02531}
}

@misc{tang2019distillingtaskspecificknowledgebert,
  title         = {Distilling Task-Specific Knowledge from BERT into Simple Neural Networks},
  author        = {Raphael Tang and Yao Lu and Linqing Liu and Lili Mou and Olga Vechtomova and Jimmy Lin},
  year          = {2019},
  eprint        = {1903.12136},
  archiveprefix = {arXiv},
  primaryclass  = {cs.CL},
  url           = {https://arxiv.org/abs/1903.12136}
}

@article{Gilardi_2023,
  title     = {ChatGPT outperforms crowd workers for text-annotation tasks},
  volume    = {120},
  issn      = {1091-6490},
  url       = {http://dx.doi.org/10.1073/pnas.2305016120},
  doi       = {10.1073/pnas.2305016120},
  number    = {30},
  journal   = {Proceedings of the National Academy of Sciences},
  publisher = {National Academy of Sciences},
  author    = {Gilardi, Fabrizio and Alizadeh, Meysam and Kubli, Maël},
  year      = {2023},
  month     = jul
}

@inproceedings{Qin_2018,
  series     = {SIGIR ’18},
  title      = {Enhancing Person-Job Fit for Talent Recruitment: An Ability-aware Neural Network Approach},
  url        = {http://dx.doi.org/10.1145/3209978.3210025},
  doi        = {10.1145/3209978.3210025},
  booktitle  = {The 41st International ACM SIGIR Conference on Research \& Development in Information Retrieval},
  publisher  = {ACM},
  author     = {Qin, Chuan and Zhu, Hengshu and Xu, Tong and Zhu, Chen and Jiang, Liang and Chen, Enhong and Xiong, Hui},
  year       = {2018},
  month      = jun,
  pages      = {25–34},
  collection = {SIGIR ’18}
}

@inproceedings{bian2020learningmatchjobsresumes,
  title={Learning to Match Jobs with Resumes from Sparse Interaction Data using Multi-View Co-Teaching Network},
  author={Bian, Shuqing and Chen, Xu and Zhao, Wayne Xin and Zhou, Kun and Hou, Yupeng and Song, Yang and Zhang, Tao and Wen, Ji-Rong},
  booktitle={Proceedings of the 29th ACM International Conference on Information \& Knowledge Management},
  pages={65--74},
  address={New York, NY, USA},
  publisher={Association for Computing Machinery},
  year={2020}
}

@inproceedings{10.1145/3357384.3357949,
  author    = {Le, Ran and Hu, Wenpeng and Song, Yang and Zhang, Tao and Zhao, Dongyan and Yan, Rui},
  title     = {Towards Effective and Interpretable Person-Job Fitting},
  year      = {2019},
  isbn      = {9781450369763},
  publisher = {Association for Computing Machinery},
  address   = {New York, NY, USA},
  url       = {https://doi.org/10.1145/3357384.3357949},
  doi       = {10.1145/3357384.3357949},
  booktitle = {Proceedings of the 28th ACM International Conference on Information and Knowledge Management},
  pages     = {1883–1892},
  numpages  = {10},
  location  = {Beijing, China},
  series    = {CIKM '19}
}

@misc{jouanneau2026efficientlongcontextrankingarchitecture,
  title         = {An Efficient Long-Context Ranking Architecture With Calibrated LLM Distillation: Application to Person-Job Fit},
  author        = {Warren Jouanneau and Emma Jouffroy and Marc Palyart},
  year          = {2026},
  eprint        = {2601.10321},
  archiveprefix = {arXiv},
  primaryclass  = {cs.CL},
  url           = {https://arxiv.org/abs/2601.10321}
}

@misc{fu2025lanternscalabledistillationlarge,
  title         = {LANTERN: Scalable Distillation of Large Language Models for Job-Person Fit and Explanation},
  author        = {Zhoutong Fu and Yihan Cao and Yi-Lin Chen and Aman Lunia and Liming Dong and Neha Saraf and Ruijie Jiang and Yun Dai and Qingquan Song and Tan Wang and Guoyao Li and Derek Koh and Haichao Wei and Zhipeng Wang and Aman Gupta and Chengming Jiang and Jianqiang Shen and Liangjie Hong and Wenjing Zhang},
  year          = {2025},
  eprint        = {2510.05490},
  archiveprefix = {arXiv},
  primaryclass  = {cs.CL},
  url           = {https://arxiv.org/abs/2510.05490}
}

@misc{yu2026confitv3improvingresumejob,
  title         = {ConFit v3: Improving Resume-Job Matching with LLM-based Re-Ranking},
  author        = {Xiao Yu and Ruize Xu and Chengyuan Xue and Junyu Chen and Matthew So and Shijun Ma and Bo Liu and Xiangye Liang and Zhou Yu},
  year          = {2026},
  eprint        = {2605.09760},
  archiveprefix = {arXiv},
  primaryclass  = {cs.CL},
  url           = {https://arxiv.org/abs/2605.09760}
}

@inproceedings{chen2017enhanced,
  title={Enhanced {LSTM} for Natural Language Inference},
  author={Chen, Qian and Zhu, Xiaodan and Ling, Zhen-Hua and Wei, Si and Jiang, Hui and Inkpen, Diana},
  booktitle={Proceedings of the 55th Annual Meeting of the Association for Computational Linguistics (Volume 1: Long Papers)},
  pages={1657--1668},
  address={Vancouver, Canada},
  publisher={Association for Computational Linguistics},
  year={2017}
}

@inproceedings{conneau2017supervised,
  title={Supervised Learning of Universal Sentence Representations from Natural Language Inference Data},
  author={Conneau, Alexis and Kiela, Douwe and Schwenk, Holger and Barrault, Lo\"ic and Bordes, Antoine},
  booktitle={Proceedings of the 2017 Conference on Empirical Methods in Natural Language Processing},
  pages={670--680},
  address={Copenhagen, Denmark},
  publisher={Association for Computational Linguistics},
  year={2017}
}

@inproceedings{mou2016natural,
  title={Natural Language Inference by Tree-Based Convolution and Heuristic Matching},
  author={Mou, Lili and Men, Rui and Li, Ge and Xu, Yan and Zhang, Lu and Yan, Rui and Jin, Zhi},
  booktitle={Proceedings of the 54th Annual Meeting of the Association for Computational Linguistics (Volume 2: Short Papers)},
  pages={130--136},
  address={Berlin, Germany},
  publisher={Association for Computational Linguistics},
  year={2016}
}

@inproceedings{thakur2021augmentedsbert,
  title={Augmented {SBERT}: Data Augmentation Method for Improving Bi-Encoders for Pairwise Sentence Scoring Tasks},
  author={Thakur, Nandan and Reimers, Nils and Daxenberger, Johannes and Gurevych, Iryna},
  booktitle={Proceedings of the 2021 Conference of the North American Chapter of the Association for Computational Linguistics: Human Language Technologies},
  pages={296--310},
  address={Online},
  publisher={Association for Computational Linguistics},
  year={2021}
}

\appendix

\section{Comparison with Closest Work}
\label{app:related}

Table~\ref{tab:related} summarizes how our approach differs from the
closest LLM-for-person--job-fit systems discussed in
\S\ref{sec:positioning}.

\begin{table*}[t]
\centering\small
\setlength{\tabcolsep}{4pt}
\caption{Qualitative comparison with the closest LLM-for-person--job-fit
  systems. Prior systems distil or prompt an LLM into a single score (or
  a free-text rationale); ours exposes eight independently supervised,
  named dimensions.}
\label{tab:related}
\begin{tabularx}{\textwidth}{@{}lll>{\raggedright\arraybackslash\hyphenpenalty=10000\exhyphenpenalty=10000}Xl@{}}
\toprule
\textbf{Work} & \textbf{Student model} & \textbf{Supervision} &
  \textbf{Interpretable output} & \textbf{Serving} \\
\midrule
\citet{li-etal-2025-enhancing-talent}         & Role-aware MoE ranker      & LLM-generated job descr.        & Single ranking score                        & GPU \\
\citet{yu2026confitv3improvingresumejob}           & LLM re-ranker (Qwen3)      & Contrastive + LLM rerank        & Single relevance score                      & GPU \\
\citet{fu2025lanternscalabledistillationlarge}              & Distilled generative LLM   & LLM (fit + explanation)         & Free-text explanation                       & GPU \\
\citet{jouanneau2026efficientlongcontextrankingarchitecture} & Late cross-attn.\ ranker   & Generative LLM, calib.\ KD      & Single skill-fit score                      & GPU \\
\midrule
\textbf{Ours}                    & Bi-encoder (0.6B+LoRA)     & Prod.\ LLM $\to$ cross-enc.\ KD & Eight named ordinal/binary dimensions & \textbf{CPU} \\
\bottomrule
\end{tabularx}
\end{table*}

\section{Feature Definitions and Label Schemes}
\label{app:features}

Table~\ref{tab:app_features} lists the eight matching dimensions
predicted by the feature bi-encoder.
Seven dimensions use ordinal modelling with three ordered classes
(score 0--2); certifications uses binary prediction after the
vacancy-only applicability gate removes the \texttt{not\_required}
case (Section~\ref{sec:applicability}).
\begin{table*}[tp]
\centering\small
\renewcommand{\arraystretch}{1.25}
\begin{tabularx}{\textwidth}{l>{\raggedright\arraybackslash\hyphenpenalty=10000\exhyphenpenalty=10000}p{4.2cm}X}
\toprule
\textbf{Feature} & \textbf{Classes (0 $\to$ 1 $\to$ 2)} &
  \textbf{What is evaluated} \\
\midrule
\multicolumn{3}{l}{\textit{Skill group (evaluated jointly by combined-skill prompt)}} \\[3pt]
Skill Context
  & no\_overlap $\to$ research $\to$ production
  & Context in which the overlapping required skill was applied:
    no evidence, R\&D\,/\,experimentation, or live
    production\,/\,business delivery. \\[3pt]
Experience Recency
  & old $\to$ recent $\to$ active
  & How recently the candidate used the main required skill cluster
    ($>$3 yr, 1--3 yr, or current\,/\,$<$1 yr). \\[3pt]
Skill Proficiency
  & no\_overlap $\to$ practical $\to$ professional
  & Depth of hands-on ownership: no credible evidence,
    real applied use with limited scope, or substantial ownership and
    repeated delivery. \\[3pt]
Hard Skill Experience
  & mismatch $\to$ close\_match $\to$ match
  & How well the demonstrated hands-on hard-skill experience (type
    and level) meets the vacancy requirement. \\[6pt]
\multicolumn{3}{l}{\textit{Fit group (separate per-dimension prompts)}} \\[3pt]
Industry Experience
  & not\_connected $\to$ adjacent $\to$ direct
  & Alignment of the candidate's industry background with the
    vacancy domain; most-recent role weighted most heavily. \\[3pt]
Seniority Alignment
  & misaligned $\to$ partial $\to$ aligned
  & Match between the candidate's professional level and the
    vacancy's seniority requirement. \\[3pt]
Job Title Match
  & mismatch $\to$ adjacent $\to$ match
  & Overlap between the candidate's role history (seniority markers
    stripped) and the target job title. \\[6pt]
\multicolumn{3}{l}{\textit{Requirements group}} \\[3pt]
Certifications (binary)
  & missing $\to$ met
  & Whether the candidate holds the required professional
    certifications, licences, or clearances.
    Evaluated only when the vacancy-only applicability head flags
    certification as required (Section~\ref{sec:applicability}). \\
\bottomrule
\end{tabularx}
\caption{The eight matching dimensions predicted by the feature
  bi-encoder.
  Class~2 always represents the strongest match.
  Certifications is treated as binary after applicability gating;
  the remaining seven dimensions use three-class ordinal modelling
  via cumulative logits.}
\label{tab:app_features}
\end{table*}

\section{LLM Prompt Templates}
\label{app:prompts}

All eight features are evaluated with production-refined prompts
served from a YAML prompt registry.
A shared system context (Figure~\ref{lst:shared}) injects the
vacancy and resume into every call.
The four skill-related features are evaluated jointly in a single
LLM call (Figure~\ref{lst:combined}) to ensure consistency across
correlated dimensions; the remaining four features (industry
experience, seniority alignment, job-title match, certifications)
use analogous single-feature prompts with the same structured JSON
output format.
A separate vacancy-only prompt (the \emph{job requirement gate})
determines whether education, certifications, and languages are
meaningfully required by the vacancy before any pairwise evaluation
is performed.

Company and user identifiers are omitted; all templates were
deployed verbatim in the production system and used for offline
labelling of the 168{,}772 training pairs.

\subsection{Prompt Refinement Protocol}
\label{app:prompt_protocol}

For three weeks after pilot launch, the ML team manually inspected
recruiter-recorded corrections and grouped recurring errors into failure
patterns; recruiters did not edit prompts directly. The team translated
each pattern into a generalizable rule and tested revisions against 55
new error cases and a fixed regression set of 192 human-reviewed feature
values (247 values total). Exactly three prompt updates were released,
for Skill Context, Skill Proficiency, and Certifications. A revision had
to fix the target failure, generalize to related cases, produce an
appropriate explanation, and cause no regression on the fixed set.
These values supported iterative prompt development and are not an
independent test set.

\begin{figure}[h]
\footnotesize
\begin{verbatim}
[SYSTEM — shared context, all evaluators]

You are an expert technical recruiter
and HR specialist. Your task is to
evaluate a candidate's fit for a specific
job opening.

You will be provided with the JOB POSTING
and the CANDIDATE RESUME. Subsequently,
you will receive specific feature
evaluation instructions.

<job_posting>
{job_description}
</job_posting>

<resume>
{candidate_text}
</resume>
\end{verbatim}
\caption{Shared system context prepended to every LLM
  call. Vacancy and resume are filled at runtime.}
\label{lst:shared}
\end{figure}

\begin{figure*}[h]
\small
\begin{verbatim}
[SYSTEM — combined skill evaluator]

You are a senior technical recruiter evaluating four related
hard-skill features for the same job-resume pair in one request.

Evaluate these four outputs together against the same main
overlapping required hard skill or required hard-skill cluster:
skill_context, experience_recency, skill_proficiency,
hard_skill_experience.

Core rules:
- First identify the main required hard skill or small required
  skill cluster in the job.
- Decide whether the resume shows credible evidence for that same
  skill or a very close equivalent.
- Weight the most recent / current role most heavily.
- Keep all four outputs aligned to the same overlap target, but
  do not let one field mechanically determine the others.
- Prefer conservative labels when evidence is mixed or partial.

Feature definitions:
- skill_context: production | research | academic | no_overlap
  production = skill used in live business/product workflows.
  research   = skill used in R&D / experimentation.
  academic   = coursework, thesis, or personal projects only.
  no_overlap = insufficient credible overlap to assign a context.
- experience_recency: active | recent | old
  active = current or <1 yr.  recent = 1-3 yr.  old = >3 yr.
- skill_proficiency: professional | practical | studied | no_overlap
  professional = substantial ownership and repeated delivery.
  practical    = real applied use, lighter depth or scope.
  studied      = learned / assisted / lightly practiced.
  no_overlap   = insufficient credible overlap to assess.
- hard_skill_experience: match | close_match | mismatch
  How well demonstrated hands-on experience meets the requirement.

Return valid JSON only. For each feature: score (exact label) +
reason (one sentence).

[USER]

Evaluate all four hard-skill features in one pass.

1. Identify the main required hard skill / skill cluster.
2. Check whether the resume shows real evidence for that skill
   or a close equivalent.
3. Use the same overlap target for all four outputs.
4. Be evidence-based and conservative.

Output valid JSON matching exactly this schema:
{
  "skill_context":
      {"score": "<production|research|academic|no_overlap>",
       "reason": "<1 sentence>"},
  "experience_recency":
      {"score": "<active|recent|old>",
       "reason": "<1 sentence>"},
  "skill_proficiency":
      {"score": "<professional|practical|studied|no_overlap>",
       "reason": "<1 sentence>"},
  "hard_skill_experience":
      {"score": "<match|close_match|mismatch>",
       "reason": "<1 sentence>"}
}
\end{verbatim}
\vspace{-9pt}
\caption{Combined skill evaluator prompt (anonymized).
  This single LLM call evaluates four correlated skill dimensions
  jointly, anchoring all outputs to the same required skill cluster.
  Separate single-feature prompts with the same JSON format are
  used for industry experience, seniority alignment, job-title
  match, and certifications.
  In our experiments, the \texttt{research} and \texttt{academic}
  classes of \texttt{skill\_context} were merged into a single class
  due to limited annotated examples and their close semantic meaning
  (both reflect non-production use of a skill).}
\label{lst:combined}
\end{figure*}

\section{Stability and Agreement Protocols}
\label{app:stability}

\textbf{Repeated-generation policy.}
On a stratified sample of 5{,}000 pairs, each feature label was generated
five times. Unanimous five-run agreement ranged from 86.7\% to 98.7\%
across dimensions, while first-run agreement with the five-run majority
ranged from 94.8\% to 99.7\%. Job Title Match, Experience Recency, Skill
Context, Skill Proficiency, Hard Skill Experience, and Certifications
each reached at least 98\% first-run/majority agreement and use one run
at scale. Industry Experience (95.9\%) and Seniority Alignment (94.8\%)
use two initial runs, a third after disagreement, and up to five for
ambiguity. Of three cases without a stable majority, two corrupted or
nearly empty resumes were excluded and one genuinely ambiguous case was
manually reviewed. These measurements describe run-to-run stability,
not calibrated probabilistic uncertainty.

\textbf{Agreement intervals.}
Wilson score intervals in Table~\ref{tab:feedback} use the exact
agreement count for each dimension. To account for dependence among
feature values from the same production entity, we additionally
resampled the 114 unique candidates and, separately, the 45 vacancies
with replacement for 10{,}000 iterations, recomputing pooled exact
agreement each time. The resulting candidate-cluster and
vacancy-cluster percentile intervals are 94.67--96.89\% and
94.11--97.37\%, respectively. These intervals address clustering in
the selected feedback subset but do not make that subset representative
of all surfaced candidates.

\section{Per-Feature Evaluation Results}
\label{app:per_feature}

Table~\ref{tab:app_perfeature} reports macro-F1 per feature and per
experimental condition, corresponding to the aggregated results in
Table~\ref{tab:feature_results} of the main paper.
Each cell reports performance on the same held-out, vacancy-disjoint
validation split; the \emph{Avg} row is the macro-average across all eight
features.

\begin{table}[h]
\centering\small
\setlength{\tabcolsep}{5pt}
\begin{tabular}{lccc}
\toprule
\textbf{Feature}
  & \textbf{Majority} & \textbf{MLP,}
  & \textbf{Full} \\
  & \textbf{baseline} & \textbf{orig.\ enc.} & \textbf{model} \\
\midrule
Skill Context         & 0.489 & 0.788 & \textbf{0.883} \\
Skill Proficiency     & 0.240 & 0.642 & \textbf{0.801} \\
Hard Skill Experience & 0.218 & 0.739 & \textbf{0.846} \\
Industry Experience   & 0.203 & 0.682 & \textbf{0.853} \\
Experience Recency    & 0.279 & 0.467 & \textbf{0.828} \\
Job Title Match       & 0.220 & 0.726 & \textbf{0.928} \\
Certifications        & 0.236 & 0.699 & \textbf{0.857} \\
Seniority Alignment   & 0.262 & 0.612 & \textbf{0.795} \\
\midrule
\textbf{Avg}          & 0.268 & 0.669 & \textbf{0.849} \\
\bottomrule
\end{tabular}
\caption{Macro-F1 per feature on the held-out, vacancy-disjoint
  validation split.
  \textbf{Majority baseline}: majority-class prediction.
  \textbf{MLP, orig.\ enc.}: MLP head on original (non-fine-tuned)
  encoder embeddings.
  \textbf{Full model}: fine-tuned encoder with knowledge distillation,
  monotonicity, and margin losses (trained with the full hybrid loss).
  Bold marks the best result per feature.}
\label{tab:app_perfeature}
\end{table}

\section{Training and Implementation Details}
\label{app:impl}

\paragraph{Encoder.}
We use \texttt{Qwen3-Embedding-0.6B} adapted with LoRA
($r{=}32$, $\alpha{=}64$, dropout~0.1) applied to all attention
projection layers (\texttt{q}, \texttt{k}, \texttt{v}, \texttt{o})
and all MLP gate/up/down projections.
Input sequences are truncated to 4096 tokens; representations are
taken from the last-token position; BF16 mixed precision is used
throughout.

\paragraph{MLP heads.}
Each of the eight task-specific heads processes the
4096-dimensional pair representation $\bphi$ through three linear
layers with intermediate dimensions 1024 and 512, each followed
by LayerNorm and GELU; dropout~0.1 is applied after each activation.
The certification applicability head uses a two-layer trunk
($1024{\to}768{\to}384$, LayerNorm + ReLU) followed by a linear
classifier.

\paragraph{Training.}
We train with AdamW, learning rate $10^{-4}$, weight decay $0.01$,
linear warmup over 100 steps, and a linear decay schedule.
Effective batch size is 32 (4 samples per device, gradient
accumulation over 8 steps), trained for 5 epochs.
Loss hyperparameters: $\alpha{=}0.6$ (KD weight),
$T{=}3.0$ (KD temperature), label smoothing $0.1$,
$\lambda_m{=}0.05$ (monotonicity), $\lambda_a{=}0.1$
(margin).
Stage~1 freezes the LoRA parameters and trains only the MLP heads;
Stage~2 unfreezes all LoRA parameters and fine-tunes jointly.

\paragraph{Cross-encoder teacher.}
The intermediate cross-encoder (\texttt{Qwen3-4B} reranker) is
trained on the LLM hard labels with the same hybrid loss
(minus the KD term).
Its per-pair logits are cached before bi-encoder training begins.

\paragraph{Hardware.}
All experiments were run on a single NVIDIA A100-80GB GPU.
Stage~1 completes in approximately 1.5 hours; Stage~2 in
approximately 3 hours.
The deployed model runs inference on commodity x86 CPU instances
(4 vCPUs and 8\,GiB RAM each, autoscaled), with resume embeddings
precomputed and cached. Embedding precompute is performed offline on a
single NVIDIA~T4 instance (not the A100), and serving itself
uses no GPU.

\end{document}